\documentclass[runningheads,a4paper]{llncs}

\usepackage{amssymb}
\usepackage[english]{babel}
\usepackage{inputenc}
\usepackage{algorithm}
\usepackage{amsmath}

\usepackage[noend]{algpseudocode}
\usepackage{caption}
\usepackage{graphicx}
\usepackage{array}
\usepackage{url}

\usepackage{tabularx}

\usepackage{lipsum}
\usepackage{stfloats}
\usepackage{graphicx}
\usepackage{subfig}
\usepackage{enumitem}
\usepackage{color,soul}
\usepackage{multirow}

\urldef{\mailsa}\path|tooba.aamir@rmit.edu.au|    

\algnewcommand\And{\textbf{AND }}

\begin{document}

\title{Heuristics based Mosaic of Social-Sensor Services for Scene Reconstruction}

\author{Tooba Aamir\inst{1}, Hai Dong\inst{1}\thanks{Corresponding author.
This paper has been accepted by The 21st International Conference on Web Information
Systems Engineering (WISE 2020). Please cite the paper as
\textbf{Aamir, T., Dong, H., Bouguettaya, A.: Heuristics based Mosaic of Social-Sensor Services for Scene Reconstruction. The 21st International Conference on Web Information Systems Engineering (WISE 2020) (2020)}
} and Athman Bouguettaya\inst{2}}

\institute{School of Science, RMIT University, Melbourne, Australia
\\
\email{\{tooba.aamir,hai.dong\}@rmit.edu.au}
\and
School of Computer Science, The University of Sydney, Sydney, Australia
\\
\email{athman.bouguettaya@sydney.edu.au}
}
 
\authorrunning{Tooba Aamir, Hai Dong and Athman Bouguettaya }

\maketitle

\begin{abstract}

We propose a heuristics-based social-sensor cloud service selection and composition model to reconstruct \textit{mosaic} scenes. The proposed approach leverages \textit{crowdsourced} social media images to create an image mosaic to reconstruct a scene at a designated location and an interval of time. The novel approach relies on the set of features defined on the bases of the image metadata to determine the \textit{relevance} and \textit{composability} of services. Novel heuristics are developed to filter out non-relevant services. Multiple machine learning strategies are employed to produce \textit{smooth} service composition resulting in a mosaic of relevant images indexed by geolocation and time. The preliminary analytical results prove the feasibility of the proposed composition model.

\end{abstract}

\section{Introduction}

Advancements in smart devices, e.g., smartphones, and prevalence of social media, have established new means for information sensing and sharing [1][2]. Social media data, e.g., Twitter posts and Facebook statuses, have become a significant and accessible means for sharing the facts or opinions about any event. Crowdsourcing (i.e., sensing, gathering and sharing) of social media data is termed as \textit{social sensing} [1][2]. Smart devices (i.e., \textit{social-sensors} [1]) have the ability to embed sensed data directly into social media outlets (i.e., \textit{social clouds} or \textit{social-sensor clouds}) [4]. Monitoring the social media data (i.e., \textit{social-sensor data}) provides multiple benefits in various domains. For example, traditional sensors like CCTVs are usually unable to provide complete coverage due to their limited field of view and sparsity. Social-sensor data, especially multimedia content (e.g., images) and related metadata may provide a multi-perspective coverage [1]. We focus on utilising offline social-sensor images, i.e., downloaded batches of social media images and their related metadata to assist in scene reconstruction. \par
 
The social-sensor images are featured with diverse formats, structures and sources. There are various inherent challenges for the efficient management and delivery of such multifaceted social-sensor images [3][4][7]. One of the significant challenges is \textit{searching and analysing a massive amount of heterogeneous data} in the inter-operable social cloud platforms [3]. \textit{Service paradigm} is proposed to address this challenge. Service paradigm helps to convert social-sensor data into simplified and meaningful information, by abstracting away the data complexity [4][5]. Service paradigm abstracts a social-sensor image (i.e., an incident-related image posted on social media) as \textit{an independent function}, namely \textit{a social-sensor cloud service} (abbreviated as a SocSen service). A SocSen service has \textit{functional}  and \textit{non-functional} properties. The functional attributes of the camera include taking images, videos and panoramas, etc. The spatio-temporal and contextual information of a social-sensor image is presented as the \textit{non-functional properties} of a service. This abstraction allows a uniform and efficient delivery of social-sensor images as a SocSen service, without relying on \textit{image processing} [4][5]. \par 

SocSen services provide efficient access to social media images for scene reconstruction and make them easy to reuse in multiple applications over diverse platforms. However, the challenge is designing an efficient selection and composition solution for context-relevant SocSen services based scene reconstruction. Context relevance includes covering the same incident or segment of an area in a given time. The existing techniques developed for SocSen service selection and composition are primarily dependent upon the prerequisite that objects in social media images can be explicitly identified and/or explicit definitions of relevance and composability are provided [4][5][6][12][13]. For example, two cars in two images can be automatically identified as identical by social media platforms. This type of object identification technologies is currently \textit{unavailable} in most of the mainstream social media platforms. Moreover, the existing approaches analyse the spatio-temporal aspects of services and their relevance to queries as per the explicitly defined rules to achieve a successful service composition for scene reconstruction [4][5][12][13]. Defining these rules needs substantial human \textit{intervention} and rely heavily on \textit{domain experts' knowledge} in specific scenes. This dependency limits the generality of these approaches. 

This paper proposes to employ \textit{heuristics and machine learning to automatically select relevant and composable SocSen services based on the services' features and our designed heuristics}. The proposed approach provides a strategy to enable automatic SocSen service selection and composition for analysing scenes \textit{without using image processing and objects identification technologies}. The proposed composition approach will form scenes by selecting spatially close and relevant images and placing them in a smooth mosaic-like structure. \par
The major contributions of our proposed work are:
\begin{itemize}
\item We design an approach enabling automatic and smooth mosaic scene construction based on social-sensor images, catering to most social media platforms that have not fully realised image processing and object identification.
\item We explore various machine learning based classification strategies, including decision tree, support vector machine, and artificial neural network, to assess and select relevant and composable services for a composition.
\end{itemize}

\section{Motivation Scenario}
A typical scenario of area monitoring is used to illustrate the challenges in scene analysis. Let us assume, an incident occurred near the cross-section of \textit{Road X} and \textit{Road Y} during a \textit{time period t}. The surveillance of the road segment through the conventional sensors, e.g., speed cameras and CCTV, is limited. Traffic command operations require the \textit{maximum visual coverage} of the incident. The wide availability of smartphone users sharing images or posts on social networks might provide extra visual and contextual coverage. We assume that there are numerous social media images, i.e., \textit{SocSen services} available over different social clouds, e.g., Facebook, Twitter, etc.  These \textit{SocSen services} can be used to fulfil the user's need for maximum coverage. Fig. 1 shows a set of sample images taken around the queried location and time. This research proposes to \textit{leverage these social media images to reconstruct the desired scene and provide users with extra visual coverage}. The query $q = < R, TF>$ includes an approximate region of interest and the approximate time frame of the queried incident.

\begin{itemize}
    \item Query Region $R(P< x, y >, l,w)$, where $< x, y >$ is a geospatial co-ordinate set, i.e., decimal longitude-latitude position and \textit{l} and \textit{w} are the horizontal and vertical distances from the center $< x, y >$ to the edge of the region of interest, e.g., (-37.8101008,144.9634339, 10m, 10m)
    \item Query Time $TF(t_{s}, t_{e})$, where $t_{s}$ is the start time of the query, e.g., 13:20 AEST, 20/01/2017 and $t_{e}$ is the end time of the query, e.g., 14:00 AEST, 20/01/17.
\end{itemize}
The process of scene reconstruction is considered as alleviating a service selection and composition problem. This research proposes to \textit{leverage heuristics and machine learning based techniques for the selection and composition of SocSen services}. SocSen services possess multiple non-functional properties, adding value to unstructured social-sensors data. The non-functional properties of such services include (but not limited to) \textit{spatial, temporal, and angular} information of the image. The main objective of this project is to \textit{leverage heuristics and machine learning based techniques to select the services that are in the same information context, i.e., smoothly covering the area required by the user, by assessing their relevance and composability}. The composability assessment is based on \textit{machine learning strategies with features defined upon non-functional attributes of the services}. Eventually, we aim to build a SocSen service composition to fulfil the user's requirement of visual coverage. The composition is a visual summary of the queried scene in the form of a mosaic-like structure by selecting the composable services covering the area. The composed visual summary is a set of spatio-temporally similar 2D images reflecting the queried incident. The quality of the composition relies on the smoothness and the spatial continuity of the mosaic. Therefore, adequately assessed composability between services is essential for a successful composition.
\begin{figure}[h]
\centering
  \includegraphics[scale=0.60]{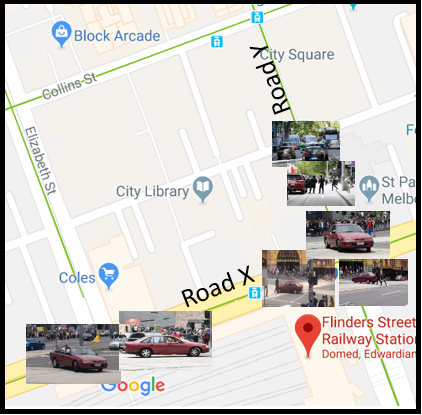}
  \caption{Motivation Scenario - A Set of Images from Social Networks}
\end{figure}
\section{Model for Social-Sensor Cloud Service}
In this section, we define concepts related to the social-sensor cloud service modelling. 

An atomic SocSen service \textit{Serv} is defined by:
\begin{itemize}
  \item \textit{Serv\_id}  is a unique service id of the service provider \textit{SocSen}. 
  \item \textit{F} is a set of functional attributes of the service \textit{Serv}.
  \item \textit{nF} is a set of non-functional properties of service \textit{Serv}.
  \end{itemize}
The functional and non-functional attributes of an atomic SocSen service are defined as:
\begin{itemize}
    \item The functional attributes of an atomic SocSen service capture the intended images, videos and/or panoramas.
    \item The non-functional attributes capture spatio-temporal and contextual behaviour depicted in an \textit{atomic service}. The following are the minimal non-functional attributes associated with an atomic service:
    \begin{itemize} 
        \item Time, \textit{t}, is the time of the service at which the image is taken. \textit{t} can either be a single time-stamp or a short interval of time ($t_{s}$,$t_{e}$).
        \item Location, \textit{L(x,y)}, is the location of a service where \textit{x} is the longitude position and \textit{y} is the latitude position of the service.
        \item Coverage \textit{Cov} of the sensor. It defines the extent to which a service covers the scene. Coverage \textit{Cov} is defined by VisD, dir, and $\alpha$.  
        \begin{itemize}
            \item Angle \textit{dir} is the orientation angle with respect to the North. It is obtained from the digital compass sensor.
            \item Angle $\alpha$ is the angular extent of the scene covered.   $\alpha$ is calculated from the camera lens property at the current zoom level. Angle $\alpha$ in degrees is calculated as: 
            \begin{equation}
            \alpha = 2*arctan\frac{d}{2f}
            \end{equation}
          
            Where, $f$ is the effective focal length of camera and \textit{d} represents the size of the camera sensor in the direction of captured scene. For example, for 35 mm sensor, i.e., 36 mm wide and 24 mm high, $d=36$ mm would be used to obtain the horizontal angle of view and $d=24$ mm for the vertical angle.
            \item Visible distance \textit{VisD} is the maximum coverage distance. 
        \end{itemize}
    \end{itemize} 
\end{itemize}

\section{Heuristics and Machine Learning for Social-Sensor Cloud Service Selection and Composition}
We present a framework enabling heuristics and machine learning-based strategies to execute the selection and composition of the relevant and composable SocSen services for reconstructing user queried scene. The query \textit{q} can be defined as \textit{ q = (R,TF)}, giving the spatio-temporal region of interest of the required services. Query \textit{q} includes: 1) \textit{$R = \{P<x,y>,l,w\}$}, where \textit{P} is a geospatial co-ordinate set, i.e., decimal longitude-latitude position and \textit{l} and \textit{w} are length and width distance from \textit{P} to the edge of region of interest, and 2) $TF = \{t_{s}, t_{e}\}$, where $t_{s}$ is the start time of the query and $t_{e}$ is the end time of the query. 

We propose a five-step heuristics and machine learning-based service selection and composition model to realise this goal. The five major steps are: \par
\begin{enumerate}
\item \textit{Service indexing and spatio-temporal filtering.} We employ 3D R-Tree to spatio-temporally index the SocSen services [4]. The indexed services that are out of the spatio-temporal bounds defined by the user query are filtered. \par
\item \textit{Features analysis and extraction.} We define a set of independent and dependent features based on the non-functional attributes of SocSen services. These features are believed to be related to the relevance and composability of SocSen services for scene analysis.\par
\item \textit{Heuristics based service filtering.} We introduce a heuristics-based algorithm to filter out non-relevant services based on angular and direction features. \par
\item \textit{Machine learning classifier.} We explore several popular machine learning models, such as Decision tree, Support Vector Machine (SVM) and Neural Networks (NN) for assessing the relevance and composability between two services from the outputs of Step 3. The classifiers determine whether or not two services are relevant and composable based on the defined features (Table 1). The services that are non-composable with any other service are deemed as non-relevant. The details of the employed models and their configuration are  discussed in Section 5.2.\par
\item \textit{SocSen service composition.} We compose the composite services based on the one-to-one service composability to form an image mosaic. 
\end{enumerate}
\begin{table*}[t]
  \caption{Features of SocSen Services}
  \begin{center}
   \label{tab:table1}
    \begin{tabular}{|l|l|l|l|l|}
    \hline
     ID & \textbf{Feature} &    \textbf{Category} & \textbf{Source} & \textbf{Description} \\
      \hline
    F1 & \textit{L(x,y)} & Independent & SocSen & Location of the service \\
    F2 & \textit{t} & Independent & SocSen & Time of the service \\
    F3 & \textit{Dir} & Independent & SocSen & Orientation angle with respect to the North \\
    F4 & $\alpha$ & Independent & SocSen & Angular extent of the scene covered \\
    F5 & \textit{VisD} & Independent & SocSen & Maximum coverage distance \\
    F6 & $L_{2}(x,y)$ & Independent & Calculated & Triangulated location of the service coverage \\
    F7 & $L_{3}(x,y)$ & Independent & Calculated & Triangulated location of the service coverage \\
    F8 & $\alpha_{O}$ & Dependent & Calculated & Angular overlap between two services \\
    \hline
    \end{tabular}
  \end{center}
\end{table*}
\subsection{Service Indexing and Spatio-temporal Filtering}

We need an efficient approach to select spatio-temporally relevant images (i.e. services), from a large number of social media images. Spatio-temporal indexing enables the efficient selection of services. We index services considering their spatio-temporal features using 3D R-tree [4]. 3D R-tree is used as an efficient spatio-temporal index to handle time and location-based queries. It is assumed that all available services are associated with a two-dimensional geo-tagged location $L(x,t)$ and time $t$. The leaf nodes of the 3D R-tree represent services. For the effective area of query \textit{q}, we define a cube shape region BR using the user-defined rectangular query area \textit{R} and start and end time of the scene, i.e., $t_{s}$ and $t_{e}$. The region \textit{BR} encloses a set of services relevant to \textit{q}. Fig. 2 illustrates the query region \textit{R} and the bounded region \textit{BR} across time $t_{s}$ to $t_{e}$. 

\subsection{SocSen Service Features Analysis and Extraction}
 To create services based on images, we extract five basic image features from social media images' metadata. These features include services' geo-tagged locations, the time when photographs were captured, camera direction $\overrightarrow{dir}$, the maximum visible distance of the image \textit{VisD} and viewable angle $\alpha$. These features are abstracted as the non-functional properties of a service. Utilising these non-functional properties, we define several features. The features include both independent and dependent features. The former consists of the triangulation features, i.e., geo-coordinates to form an approximate triangulated field of view (FOV). The latter includes angular overlap $\alpha_{O}$ between two images.
These additional features are formulated as: \par
\begin{itemize}
\item The $FOV$ of an image, i.e., the spatial coverage area of the service can be approximated as a triangle. We also use the trigonometry to calculate the field of view as linear measurement, i.e.,\textit{ LFOV}. $LFOV$ is calculated by:
\begin{equation}
LFOV = 2 \times tan(\frac{\alpha}{2} \times VisD)
\end{equation} 
We calculate the triangular bounds of the service coverage using the triangulation rule (Equation 2 and 3) and the spherical law of cosines (Equation 4 and 5). In the spatial coverage area (Fig. 3), the service location \textit{L(x,y)} is also the location of the camera, \textit{VisD} is the maximum visible distance of the scene, and $\alpha$ is the viewable angle. The geometric parameters for the coverage triangle are calculated using:
\begin{figure}[h]
  \centering
  \begin{minipage}[b]{0.30\textwidth}
    \includegraphics[width=\textwidth]{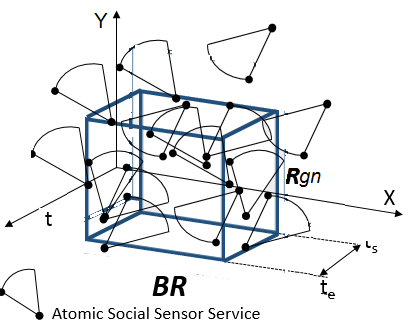}
    \caption{SocSen Service Indexing Illustration}
  \end{minipage}
 $\;$
  \begin{minipage}[b]{0.30\textwidth}
    \includegraphics [scale=0.33]{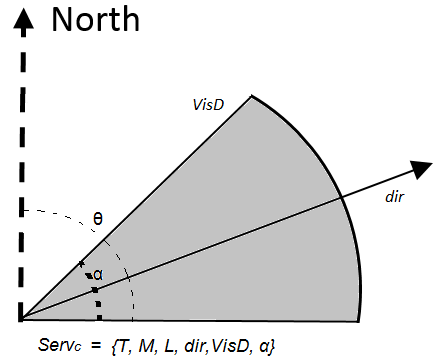}
    \caption{SocSen Service Coverage Model}
  \end{minipage}
  $\;$
  \begin{minipage}[b]{0.30\textwidth}
    \includegraphics[scale=0.50]{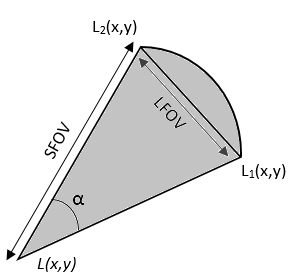}
    \caption{SocSen Service FOV Illustration}
  \end{minipage}
\end{figure}

\begin{equation}
\mid SFOV \mid = \sqrt{VisD^{2}-0.25(LFOV^{2})}
\end{equation} 
\begin{equation}
\varphi_{2} = asin( sin\varphi_{1} \cdot cos\delta +  cos\varphi_{1} \cdot sin\delta \cdot cos\theta)
\end{equation} 
\begin{equation}
\lambda_{2} = \lambda_{1} + atan2(sin\theta \cdot sin\delta \cdot cos\varphi_{1}, cos\delta-sin\varphi_{1} \cdot sin\varphi_{2})
\end{equation}

Where, $\varphi$ is latitude, $\lambda$ is longitude, $\theta$ is the bearing (clockwise from north), and $\delta$ is the angular distance d/R. $d$ is the distance travelled, and $R$ is the earth's radius. Fig. 4 shows an example of the service coverage estimated based on the FOV descriptors. \par

\item Angular overlap $\alpha_{O}$ between services $Serv_{i}$ and $Serv_{j}$ is calculated as: 
\begin{equation}
\alpha_{O}(Serv_{i},Serv_{j}) = Serv_{i}.Dir - Serv_{j}.Dir
\end{equation}
Where, $Dir$ is the orientation angle with respect to the North.
\end{itemize}

\subsection{Heuristics based Service Filtering}

We propose to filter out non-relevant services based on the direction related features in Table 1. The direction-related features rely on service coverage. The direction-related features help to assess and filter services that do not provide useful coverage of the user-required region. We propose features filtering based on a heuristic algorithm. The heuristic algorithm uses the directional and angular features of the social media images.

The classic SocSen service selection and composition techniques perform efficiently when the targeted problem completely satisfies spatial and temporal constraints. However, heuristics outperform traditional methods of composition in the noisy, sparse or discontinuous search space [9].  Heuristics have been extensively applied for feature selection, feature extraction and parameter fine-tuning.  We employ hybridisation of heuristics and machine learning to improve the performance of machine learning techniques. We use 1) parameter control strategy (for SVM and decision tree) and 2) instance-specific parameter tuning strategy for fine-tuning of the parameters.\par

Parameter control strategy employs dynamic parameter fine-tuning by controlling the parameters during the training phase. We assume that the algorithms are robust enough to provide excellent results for a set of instances of the same problem with a fixed set of parameter values.  Parameter control strategy focuses on a subset of the instances to analyse their response to the classifier.\par 

Our proposed heuristic algorithm helps to select a subset of services for classifier implementation. The algorithm uses a rank-based selection strategy, which is defined in the following steps. \par
\begin{enumerate}
\item The distance and direction difference between any two services are measured.
\item The directional relevance value (Equation 7) is evaluated for each pair of services, given distance and the directional difference value. The services are ranked from the highest to the lowest relevance.
\item All the services with relevance value above 0 are selected.
\item The distance of the services pair away the centre of the required region is measured, and the direction of the service pair with respect to the centre point is assessed (Equation 8).
\item All the service pairs are then ranked upon position relevance. The position relevance is assessed as a ratio of the direction of a service pair from the centre to the distance of the services pair from the centre. 
\end{enumerate}

To keep it simple, we chose a relatively simple form of a 2-criteria relevance function, which is defined as follows:

\begin{equation}
\begin{split}
Relevance(Serv_{i},Serv_{j}) = & Serv_{i} \cdot Serv_{j} \\ = & \parallel Serv_{i} \parallel * \parallel Serv_{j}\parallel * cos(\alpha_{O}(Serv_{i},Serv_{j}))
\end{split} \end{equation}
Where, $\parallel Serv \parallel$ is scalar magnitude of service coverage, and $Relevance$ is normalised between [-1,1]. If the services are pointing to the same direction, then relevance is positive. If the services are pointing to the opposite direction, then the relevance is negative.
\begin{equation}
Dir(Serv_{i},P) = \theta_{Serv_{i}} = cos^{-1} (\frac{(Serv_{i} \cdot P)}{\parallel Serv_{i} \parallel})
\end{equation}
Where, $Dir(Serv_{i}, P)$ is the direction of the service $Serv_{i}$ with respect to the centre point $P$ of the user's query. $P$ is a geospatial coordinate set, i.e., decimal longitude-latitude position (see Sec. 3).

\subsection{SocSen Service Composition} 
Eventually, the composition is formed by placing composable services in a mosaic-like structure. 3D R-Tree indexing helps in the spatio-temporal formation of the queried scene. The one-on-one composability is extended to deduce the composability among multiple services. This is realised by allowing transfer among multiple composable service pairs. For example, if $Serv_{a}$ is composable with $Serv_{b}$, and $Serv_{b}$ is composable with $Serv_{c}$, then $Serv_{a}$ and $Serv_{c}$ are deemed composable. The composite services comprise of a set of selected atomic services to form a visual summary of the queried scene. The visual summary is an arrangement of the 2D images, forming a mosaic-like scene. Fig. 1 shows a set of sample images taken around the same location and similar time. Fig. 5 presents the eventually composed mosaic-like scene, i.e., the composite spatio-temporal SocSen services. 

\begin{figure*}[t]
  \centering
  \includegraphics[scale=0.60]{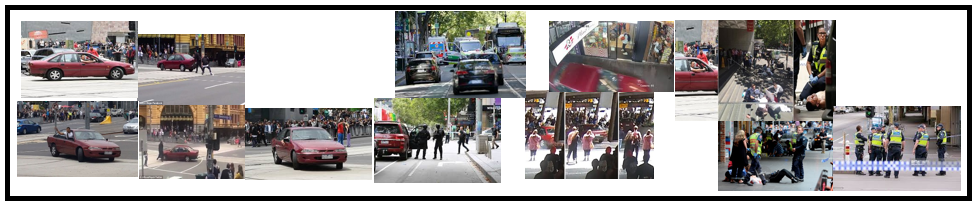}
  \caption{Mosaic-like Scene from the Sample Images}
\end{figure*} 

\section{Experiment and Evaluation}
\subsection{Experiment Setup}
Experiments are conducted to evaluate the impact of the heuristics and the performance of the classifiers on SocSen service selection and composition. To the best of our knowledge, there is no real-world spatio-temporal service dataset to evaluate our approach. Therefore, we focus on evaluating the proposed approach using a collected dataset. The dataset is a collection of 10,000 images from Jan 2016 to July 2019, in City of Melbourne, Australia. The images are downloaded from multiple social networks, e.g., flicker, twitter, google+. We retrieve and download the images related to different locations and time windows. To create services based on the images, we extract their geo-tagged locations, the time when an image was captured, and FOV information. Time, location, camera direction $\overrightarrow{dir}$, the maximum visible distance of the image \textit{VisD} and viewable angle $\alpha$ are abstracted as the non-functional property \textit{Cov} of the service. We generate 70 different queries based on the locations in our dataset. Each query \textit{q} is defined as \textit{q = (R,$t_{s}$,$t_{e}$)}, giving the region of interest and time of the query (an example can be found in Sec. V). To evaluate the service composition performance using our proposed models (SVM, Decision Tree and ANN), we conduct several experiments with the features listed in Table 1. To obtain the ground truth, we conduct a user study to gauge the perceived composability of every pair of services for each user query. All the experiments are implemented in Java and MATLAB. All the experiments are conducted on a Windows 7 desktop with a 2.40 GHz Intel Core i5 processor and 8 GB RAM.\par

\subsection{Machine Learning based Composability Assessment and Composition}
Systematic selection of the machine learning models is essential for the better performance of the proposed approach. We demonstrate the process of machine learning model selection by comparing multiple standard models with different parameter configurations. We have experimented with Decision tree, SVM and ANN for assessing the relevance and composability of services. The input data are divided into training data (70\%), validation data (15\%), and testing data (15\%). We manually identify the total numbers of services, relevant services and relevant and selected services as the ground truth for each query. Precision, recall and F1-score are employed to assess the effectiveness of all the models. \par
We opt for the decision tree due to its simple structure and co-related nature of our features. We assess the decision tree along with different depth levels and features to find the parameters producing the optimal results. We eventually chose a 20 splits based decision tree that makes a coarse distinction between composability and non-composability between two services. \par
SVM is opted due to sparse and unpredictable nature of our services' parameters [8]. The success of SVM depends on the selection of the kernel and its parameters. In our case, we experimented Quadratic kernel, Cubic kernel and Gaussian RBF kernel. All the kernels are fine-tuned to achieve the best results in terms of precision, recall and f-score. Based on the experimental results, we have selected the Quadratic kernel and Gaussian RBF kernel for further comparisons. \par
ANNs are computing systems that learn to perform tasks by considering examples (i.e., training data) generally without being programmed with task-specific rules. We opt for feed-forward multilayer ANN due to their ability to learn and infer non-linear relations among training services and further detect similar relations in unseen services. We employ ANN comprised of an input layer, multiple hidden layers (sigmoid), and an output layer (softmax) [10]. The number of hidden layers is adjusted through parameter tuning. We use all the features (i.e. F1-F8 in Table 1) to generate a vector for each service. We attempt respectively two, three and four hidden layers with ten neurons to build a feed-forward neural network. We assess our composability approach for one-to-one and many-to-one composability assessment. One-to-one composability relationship depicts composability between two neighbouring services. Many-to-one composability relationship refers to composability of a cluster of services collaboratively depicting a scene. The results of the ANN with different layers and composability relations show that the 2-layered ANN for one-to-one composability detection.

\subsection{Evaluation and Result Discussion}
We conduct the following evaluations to assess the effectiveness of the proposed approach: 1) we assess the composition performance of the machine learning models to find out the most appropriate model for service composition; 2) we compare the composition performance of the machine learning models with and without the heuristics to testify the influence of the heuristics; 3) we compare the composition performance of the models with the existing tag and object identification based composition model (abbreviated as TaOIbM) [5] and an image processing based model -  Scale-Invariant Feature Transform (SIFT) [11].
\\
\textbf{Performance of the Machine Learning Classifiers on Composition} In the first set of experiments, we analyse the performance of the tuned machine learning classifiers on service composition. We filter out services based on the dependent features. The services are then transferred to the classifiers for the relevance and composability assessment. Dependent features based filtration helps to eliminate the images with duplicated coverage and provides precise results. We assume 50\% overlap as a general heuristic fitness threshold based on the existing studies [4][5]. We use precision, recall and F1-score to assess the performance of the classifiers on the service composition. The performance of the classifiers on service composition can be found in Fig. 6. It can be observed that the ANN shows slightly better overall performance than the SVM and the decision tree. The SVM Gaussian is somewhat behind the ANN with higher precision but lower recall and F1-score.
\begin{figure}
\centering   
\begin{minipage}[b]{.3\linewidth}
    \centering
    \includegraphics[width=\textwidth]{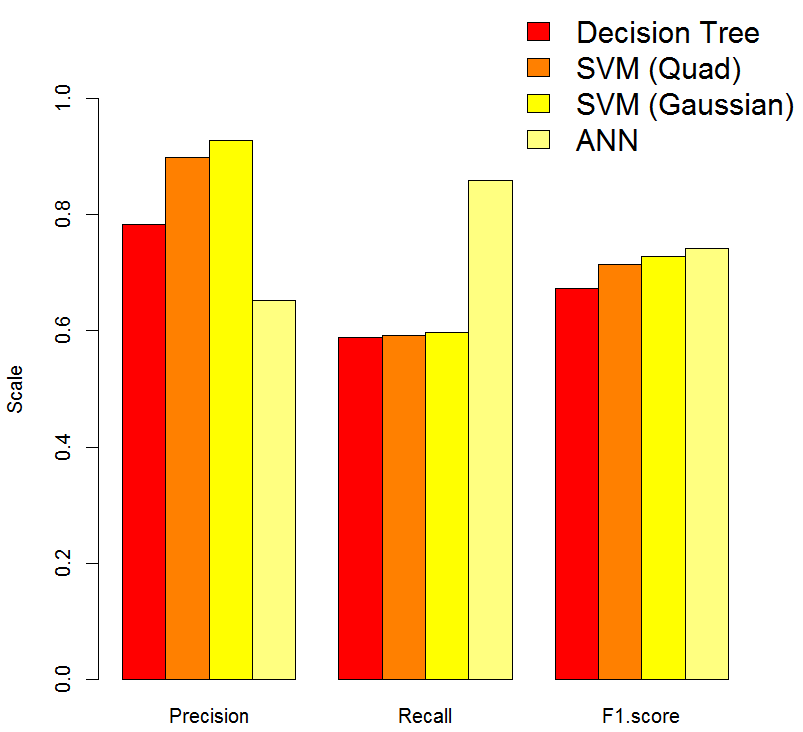}
    \caption{Classifiers with Heuristics}
\end{minipage}
$\;$
\begin{minipage}[b]{.3\linewidth}
    \centering
    \includegraphics[width=\textwidth]{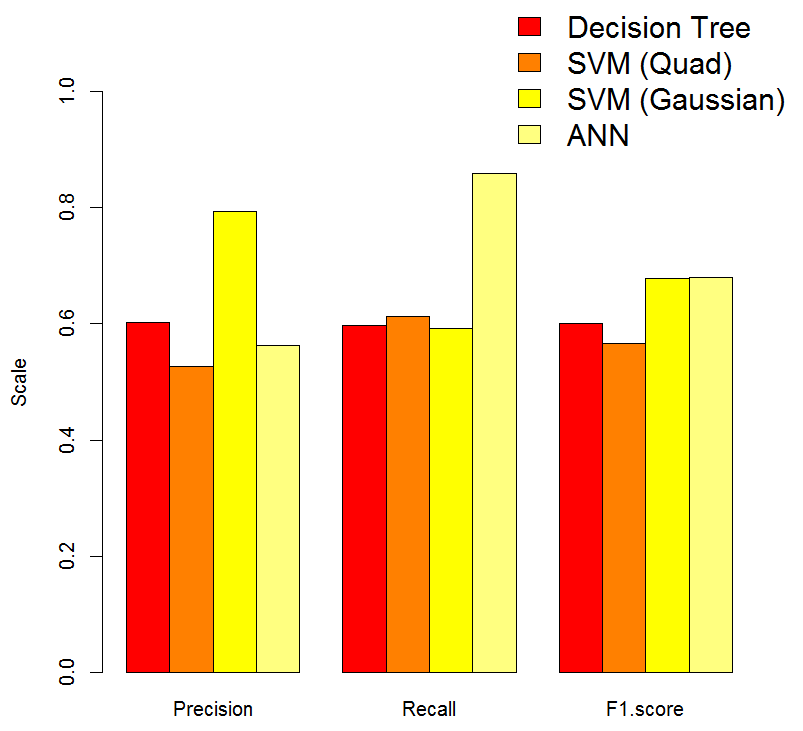}
    \caption{Classifiers without Heuristics}
  \end{minipage}
$\;$
\begin{minipage}[b]{.3\linewidth}
    \centering
    \includegraphics[width=\textwidth]{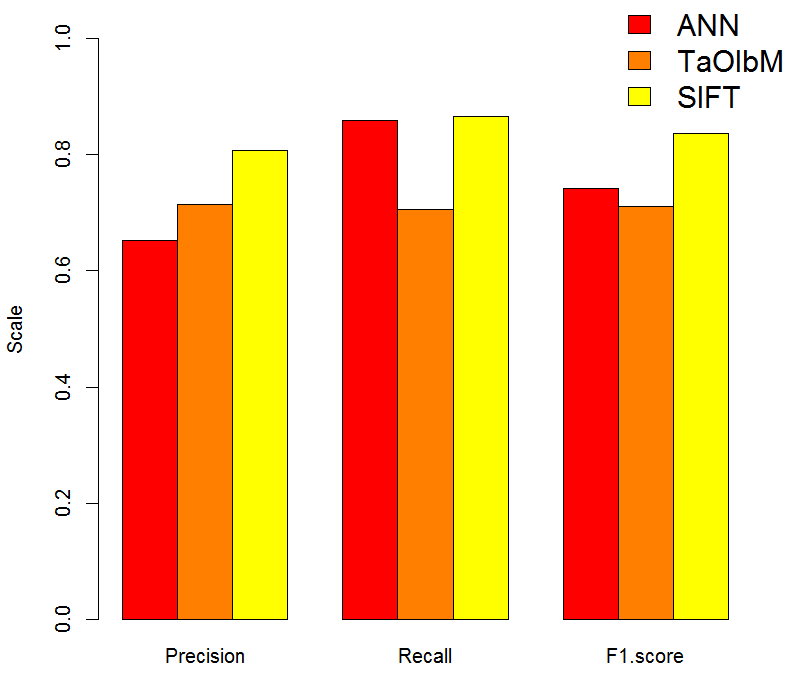}
    \caption{Comparison with TaOIbM and SIFT}
  \end{minipage}
  \end{figure}
 \\
\textbf{Impact of Heuristics on Composition} 
In the second set of experiments, we analyse the influence of the heuristics-based service filtering. We testify the performance of the machine learning classifiers with and without the heuristic algorithm. The results of the non-heuristics-based classifiers are shown in Fig. 7. It can clearly be observed that all the classifiers with heuristic outperform their counterparts. This observation validates that our heuristic algorithm is effective in filtering non-composable services. \\
\textbf{Comparison with Tag and Object Identification based Model and Image Processing Model}
The comparison result is shown in Fig. 8. First, the proposed model slightly outperforms TaOIbM. TaOIbM relies heavily on well defined composability models and tags as well as identified objects. The technologies of the latter are infeasible in most current social media platforms. Second, the proposed model has similar performance with SIFT on recall but worse performance on precision. Our model can only "guess" the image relevance and composability basing on the image metadata. In contrast, image processing models can precisely assess image relevance by processing image pixels. Therefore, it is reasonable that the precision of the former is lower than the latter. However, batch image processing is extremely time-consuming and infeasible in the enormous social media environment [2][3]. The major advantage of the proposed approach is that our proposed model does not rely on object recognition technologies and image processing.\\
\textbf{Result Discussion}
In our experiments, the heuristics are used to filter out non-composable services from the available SocSen services based on the features in Table 1. Decision tree, SVM and ANN are then used for the classification of the services passed by the heuristics into two groups: composable and non-composable services based on the features (i.e. F1-F8 in Table 1). In the experiments, the heuristic-based selection demonstrates its effectiveness in improving the performance of each classifier, by comparing the results of Fig. 6 and 7. Our model outperforms the tag and object recognition based model but has a reasonable distance with the image processing model on precision. In summary, the results of these experiments preliminarily validate the feasibility of our proposed SocSen service composition model for scene reconstruction. Especially the heuristics algorithm shows its effectiveness on improving the social media image composition performance.  
 \section{Conclusion}
This paper examines machine learning based on social-sensor cloud service selection and composition approaches. We conducted experiments to evaluate the proposed approach for accurate and smooth composition. In the future, we plan to focus on defining more effective features and exploring more machine learning models to further improve the selection and composition performance. 
 \section*{Acknowledgement}
 This research was partly made possible by DP160103595 and LE180100158 grants from the Australian Research Council. The statements made herein are solely the responsibility of the authors.

\end{document}